\documentclass[12pt]{elsarticle}

\usepackage{amssymb}
\usepackage{amsmath}
\usepackage{amsthm}
\usepackage{hyperref}
\usepackage{algorithm}
\usepackage{algorithmic}
\usepackage{xcolor}
\usepackage{adjustbox}
\usepackage{pdflscape}
\usepackage{afterpage}
\usepackage{caption}
\usepackage{subcaption}
\usepackage[normalem]{ulem}
\usepackage{setspace}

\newtheoremstyle{myplain}
  {\topsep}   
  {\topsep}   
  {\itshape}  
  {0pt}       
  {\bfseries} 
  {.}         
  {5pt plus 1pt minus 1pt} 
  {}       
\theoremstyle{myplain}

\newtheorem{lemma}{Lemma}

\theoremstyle{definition}

\newcommand{\depth}{\tt{DExp}}
\newcommand{\wad}{\tt{WAD}}
\newcommand{\waes}{\tt{WAES}}
\newcommand{\xvec}{\mathbf{x}}

\newcommand{\cvec}{\mathbf{c}}

\journal{Pattern Recognition}

\begin{document}




\title{Shallow decision trees for explainable $k$-means clustering}


\affiliation[PUC-CS]{organization={PUC-Rio -- Department of Informatics},
            city={Rio de Janeiro},
            state={RJ},
            country={Brazil}}
            
\affiliation[PUC-Math]{organization={PUC-Rio -- Department of Mathematics},
            city={Rio de Janeiro},
            state={RJ},
            country={Brazil}}

\author[PUC-CS]{Eduardo Laber}
\author[PUC-CS]{Lucas Murtinho\corref{cor1}}
\cortext[cor1]{Corresponding author.}
\ead{lucas.murtinho@gmail.com}
\author[PUC-Math]{Felipe Oliveira}

\begin{abstract}
    A number of recent works have employed decision trees for the construction of explainable partitions that aim to minimize the $k$-means cost function. These works, however, largely ignore metrics related to the depths of the leaves in the resulting tree, which is perhaps surprising considering how the explainability of a decision tree depends on these depths. To fill this gap in the literature, we propose an efficient algorithm that takes into account these metrics. In experiments on 16 datasets, our algorithm yields better results than decision-tree clustering algorithms recently presented in the literature, typically achieving lower or equivalent costs with considerably shallower trees.
\end{abstract}

\maketitle






\section{Introduction} \label{sec:intro}

As machine learning models have become used in a wide range of fields, the topic of \textit{explainability} has grown in importance. Understanding the reasoning behind a model's decision may be crucial to increase user confidence; to satisfy legal requirements; to conform to moral and ethical expectations; and to verify the model's work. Since more complex models tend to be harder to interpret but are also more capable of returning good results, there is a trade-off between model performance and explainability. The challenge of navigating this trade-off is increasingly being explored in the machine learning literature.

Although initial efforts towards explainability focused on supervised learning models \cite{DBLP:journals/jair/BurkartH21}, a number of studies on explainable unsupervised models, and clustering models in particular, have appeared more recently. One idea that has earned some attention in the literature is to partition the data based on axis-aligned cuts, which can be induced by binary decision trees: at each node $u$ of the tree, a value $\theta$ and a dimension $i$ are selected, so that all data points that have reached $u$ go to one of its two children according to whether their values for dimension $i$ are smaller than $\theta$ or not. 
In this kind of approach, usually, each cluster is associated with a leaf.

Decision trees are widely considered to be explainable models by machine learning standards. However, the explainability of a decision tree  greatly depends on the depths of its leaves, as empirically demonstrated by  \cite{DBLP:journals/eswa/PiltaverLGM16} in a study on how tree structure parameters (the number of leaves, branching factor, tree depth) influence the tree interpretability. The conclusion, based on empirical data from a survey with 98 questions answered by 69 respondents, is that the  question depth (the depth of the deepest leaf that is required when answering a question about a classification tree) turns out to be the most important parameter. Explaining leaves  that are far from the root involves many tests, which makes it harder to grasp the model's logic.

There are many possible metrics that can be associated
with the depths of the leaves, such
as the maximum depth and the average depth.
Here, we focus on metrics that consider as
equally important 
the explanation of each data point.
More specifically, we consider  
the {\em Weighted Average Depth} ($\wad$)  
and the {\em Weighted Average Explanation Size}
($\waes$).
The former weighs the depth of each leaf by the number of points of its associated cluster; to minimize it, large clusters shall be associated with shallower leaves (shorter explanations).
The latter is a variation of $\wad$
that replaces  the depth of a leaf by the 
number of non-redundant tests in the path from the 
root to the leaf. These measures are formalized and discussed more
thoroughly in Section \ref{ref:Preliminaries}.

\begin{figure*}[t]
    \centering
    \includegraphics[width=\textwidth]{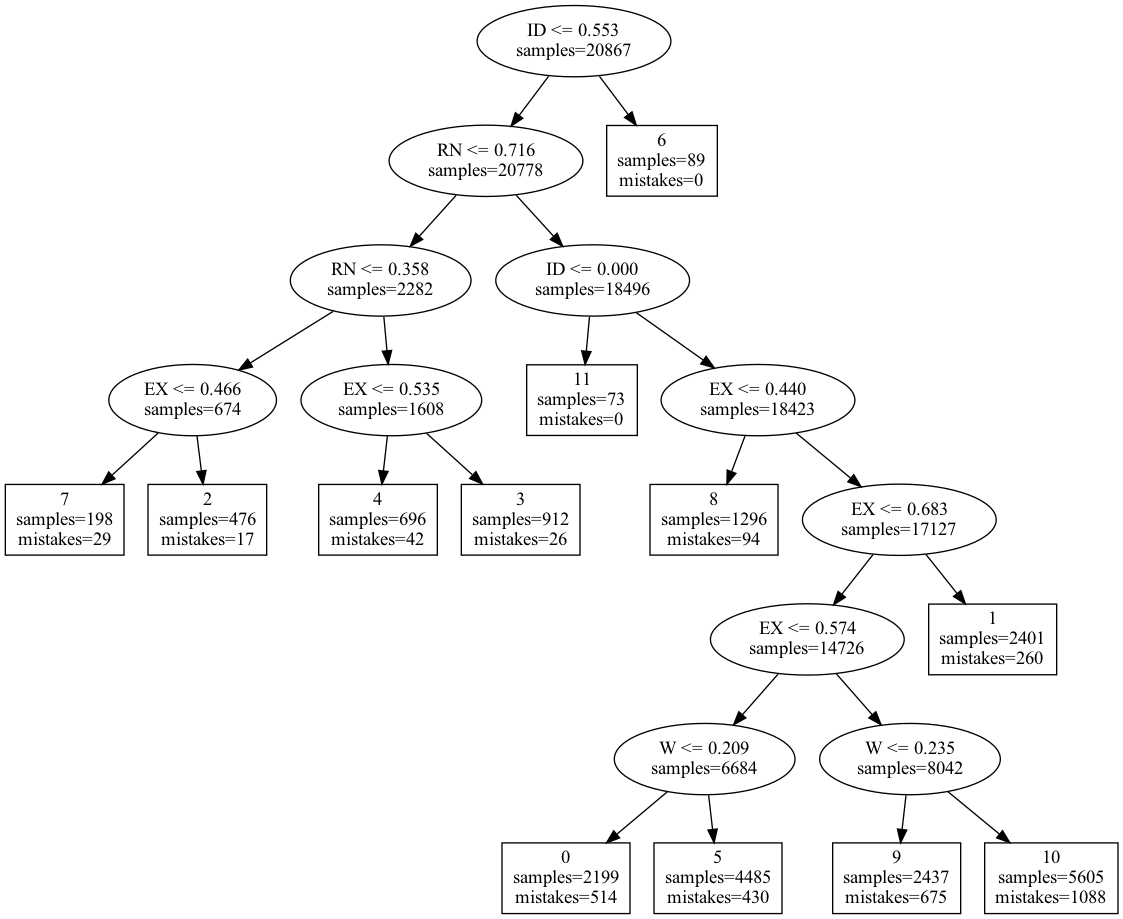}
     \caption{A tree from the {\sc ExGreedy} algorithm for the {\tt Avila} dataset. The inequalities in the elliptical nodes correspond to the condition analyzed for the samples that have reached that node. The number of mistakes indicate the number of points separated from their original centers by the preceeding cut.}
         \label{fig:exg_tree}
\end{figure*}

\begin{figure*}[t]
    \centering
    \includegraphics[width=\textwidth]{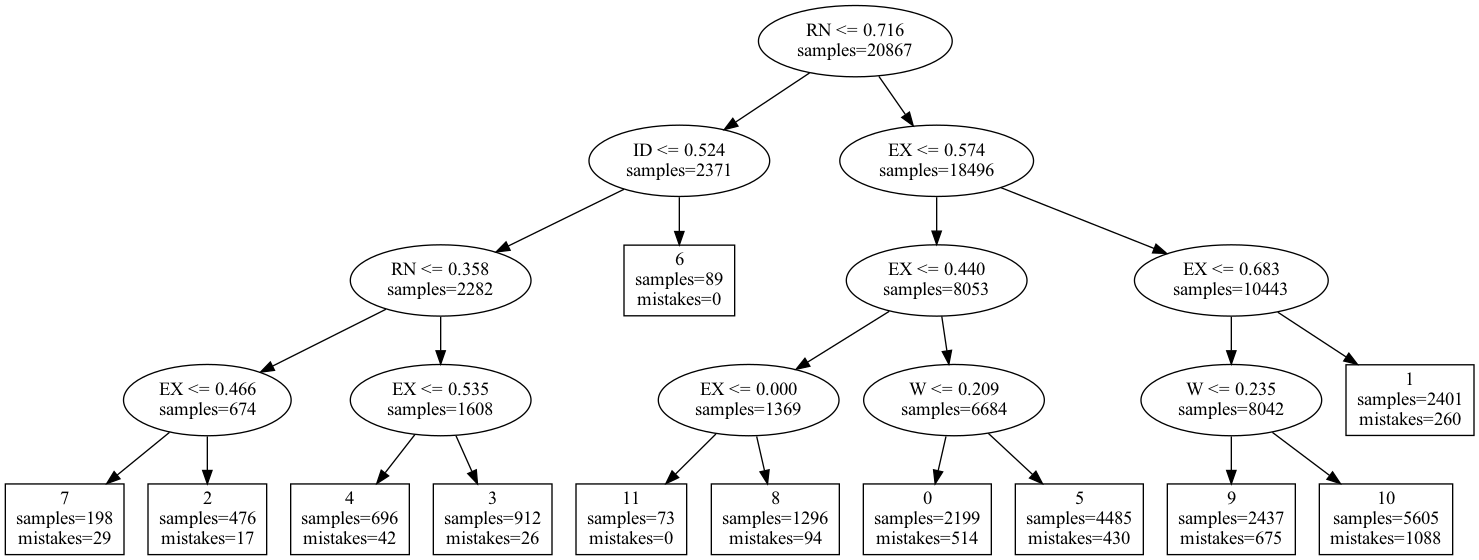}
    \caption{A tree from the {\sc ExShallow} algorithm, proposed in this paper, for the {\tt Avila} dataset. It induces the same partition as the tree presented in Figure \ref{fig:exg_tree}.}
    \label{fig:alg1_tree}
\end{figure*}

Figures \ref{fig:exg_tree} and \ref{fig:alg1_tree} show two decision trees that partition the \texttt{Avila} dataset \citep{de2018reliable} into 12 clusters. Both trees induce the same partition; however, the tree from Figure \ref{fig:exg_tree}, produced by {\sc ExGreedy} \cite{laber2021price}, has $\waes \approx$ 5.4, while the one from Figure \ref{fig:alg1_tree}  has   $\waes \approx$  3.7, which represents a ``gain'' of almost 2 conditions on average. For $\wad$, the gain is even larger, of over 2 conditions ($\approx 6.2$ vs. $\approx  3.8$). This example suggests that there is significant room to improve the explainability of the partitions provided by algorithms available in the literature.

\medskip
\noindent{\bf Our contributions.}
As in \cite{dasgupta2020explainable,frost2020exkmc}, we investigate the problem of building explainable  clustering via decision trees.  The main difference of our work with respect to the previous ones is our focus on building decision trees that simultaneously  yield short explanations  and induce  partitions of good quality in terms of the $k$-means cost function.
We understand that one  contribution of our paper is the observation that previous approaches
overlook the quite important aspect
of minimizing  measures
related to the tree's depth.

In Section \ref{sec:algos-0} we present a strategy that  builds decision trees that induce partitions of low $k$-means cost and have low values for $\waes$ and $\wad$. As other proposals in the literature, we start from the partition provided by some  algorithm for the   (non-explainable) $k$-means clustering problem and build the tree in a top-down fashion, by selecting at each node a cut that is ``good'' in terms of minimizing our metrics.  The key novelties we present here are an effective and efficient  way to evaluate the potential of a cut in terms of minimizing the $\wad / \waes$  and how to efficiently trade-off the  (potentially conflicting) goals of minimizing both these metrics and the cost of the induced partition.

To evaluate our strategy, in Section \ref{sec:experiments}, we compare its performance against recently proposed algorithms  over 16 datasets.
Our strategy generated partitions  as good as the best of its competitors in terms of the  $k$-means cost, while being significantly better
in terms of the aforementioned explainability measures. It also compares to the best of these competitors in terms of explainability, while inducing much better partitions than this competitor in terms of the $k$-means cost. Moreover, these gains were obtained without compromising  computational efficiency. 

\subsection{Preliminaries and problem definition}
\label{ref:Preliminaries}

Let ${\cal X}$ be a collection of $n$ data points in $\mathbb{R}^d$ and $k \ge 2$ be an integer. In (cost-oriented) hard clustering problems, we want to find a partition of ${\cal X}$ that minimizes a given cost function. In the widely studied $k$-means clustering problem, the cost
of a partition ${\cal P} = \{C_1,\dots,C_k\}$
is the sum of the squared Euclidean distances between all points in ${\cal X}$ and the representatives of the clusters to which they belong:
\begin{equation}
    \text{cost}({\cal P}) = \sum_{i=1}^k \sum_{\xvec \in C_i} ||\xvec - \cvec_i||^2_2. \label{eq:km_cost}
\end{equation}
In this case, the representative $\cvec_i$ of cluster $C_i$ is given by the mean of its points, 
$\cvec_i = \frac{\sum_{\xvec \in C_i} \xvec}{|C_i|}.$

In our study we are interested
in partitions
induced by axis-aligned binary decision trees.
A decision tree is axis-aligned if each internal node $v$ is associated with a
test (cut), specified by a coordinate $i_v \in [d]$ and a real value $\theta_v$, that partitions the  points
in ${\cal X}$ that reach $v$ into two sets:
those  having the  coordinate  $i_v$  smaller than or equal to $\theta_v$ and
those having  it larger than $\theta_v$. The leaves    induce a partition of  $ \mathbb{R}^d$ into axis-aligned regions and,
naturally, a partition of ${\cal X} $ into  clusters.

For our purposes,  it will be convenient to associate
a condition to each edge of the tree: the left edge leaving a  node $v$ is associated with the condition $x_{i_v} \leq \theta_{v}$ and the right  one with the condition 
$x_{i_v} > \theta_{v}$. The explanation of a cluster $C$
in a decision tree ${\cal D}$ is given by the logical
AND of the conditions associated with the edges in the
path from the root of ${\cal D}$ to the leaf associated with  $C$.  We say that a condition is {\em redundant}
with respect to cluster  $C$ if its removal 
does not change the explanation for $C$.
As an example, if the explanation
of cluster $C$ is $x_1 >30 \mbox{ AND } x_2 \leq 20 \mbox{ AND } x_1 > 70 $, then the condition 
$x_1>30$ is redundant.

We consider two explainability measures for our 
study: the Weighted Average Explaination Size ($\waes$) and the Weighted Average Depth ($\wad$). For a partition ${\cal P}=(C_1,\ldots,C_k)$ induced by a binary decision tree ${\cal D}$  with $k$ leaves, where the cluster $C_i$ is associated with the  leaf  $i$,
we have 
\begin{equation}
 {\tt WAD}({\cal D})=\frac{ \sum_{i=1}^k |C_i| \ell_i}{n}   
\label{eq:WAD}
\end{equation} 
and
\begin{equation}
 {\tt WAES}({\cal D})=\frac{ \sum_{i=1}^k |C_i| \ell_{i}^{nr}}{n},    
\label{eq:WAES}
\end{equation} 
where $l_i$ and $\ell^{nr}_i$
are, respectively,  the number of conditions and non-redundant conditions (w.r.t. $C_i$) in the path from the root to leaf $i$.
The  $\wad$ is a very natural metric and its
relevance  was advocated in
\cite{DBLP:journals/eswa/PiltaverLGM16}.
The $\waes$, to the best of our knowledge,
has not been considered before.

In terms of explainability, a decision tree is a single structure that allows us to visualize  explanations for all clusters (some of them potentially having redundant conditions), and $\wad$
gives the average length (weighted by the cluster's sizes)  of these explanations.
For each specific cluster, however, we may derive more compact explanations by removing  redundant conditions, and $\waes$ measures the average size of these explanations, again weighted by the cluster's sizes.

The problem proposed in \cite{dasgupta2020explainable}  is that of finding the  partition that minimizes (\ref{eq:km_cost}), among those that can be induced  by a decision tree of $k$ leaves. In addition to minimizing (\ref{eq:km_cost}), we also focus on building trees with low values for $\wad$ \eqref{eq:WAD} and $\waes$ \eqref{eq:WAES}.

To accomplish our goal, we note that it is important to take into account both $\wad$ and $\waes$ during the decision tree construction, since the optimization of one metric does not imply on the optimization of the other. For instance, let $X_i$ be the set of $2^{2i}-1$ points in $R^k$, where the $j$th point has all its $k$ components equal to $2^{2i}+j$. Let ${\cal X}=X_1 \cup \ldots \cup X_k$. Clearly the optimal unrestricted  $k$-partition  for   ${\cal X}$ is $(X_1,\ldots,X_k)$. This partition can be induced by many decision trees as the trees ${\cal D}_1$-${\cal D}_4$, described below, that have only one internal node per level: 
\begin{itemize}
    \item ${\cal D}_1$: the cut at level $i$
is $(i,2^{2i+1})$ so that both $\wad({\cal D}_1)$ and $\waes({\cal D}_1)$ are $\approx k$;
 \item  ${\cal D}_2$: the cut at level $i$ is $(1,2^{2(k-i)+1})$ so that both $\wad({\cal D}_2)$ and $\waes({\cal D}_2)$ are $O(1)$;
  \item  ${\cal D}_3$: the cut at level $i$ is $(i,2^{2(k-i)+1})$ so that $\wad({\cal D}_3)$ is $O(1)$ and
  $\waes({\cal D}_3)$ is $\approx k  $ 
  \item ${\cal D}_4$: the cut at level $i$ is $(1,2^{2i+1})$  so that $\wad({\cal D}_4)$ is $
  \approx k$ and
  $\waes({\cal D}_4)$ is $O(1)  $
\end{itemize}

Therefore, we shall consider both $\waes$ and $\wad$ while building the tree, otherwise we can end up with a tree that performs poorly with respect to one of the metrics.

We conclude this section by introducing  terminologies and notations that will be useful throughout this paper. We use the term {\em explainable clustering} to refer to a clustering that is induced by some axis-aligned decision tree. By an $i$-cut we mean a cut associated with component $i$, that is, a cut $x_i \leq \theta $, for some real value  $\theta$. If a node in a decision tree is   associated with an $i$-cut we say that it is an $i$-node.

\section{Related work}

\cite{dasgupta2020explainable} presents a
poly-time algorithm, \textsc{IMM}, that receives a
(non-explainable) partition ${\cal P}_u$ to the 
 $k$-means clustering problem and builds a decision tree, in 
 top-down fashion, by selecting at each node the cut that, among those that separate at least two
representatives in  ${\cal P}_u$, minimizes the number of data points separated from their representatives in ${\cal P}_u$.
In addition, they prove that  the cost of
the resulting partition
is $O(k^2) \text{cost}({\cal P}_u)$. 
A consequence of this result is that the
{\em price of explainability}, measured by the ratio between the cost of an optimal explainable partition and
that of
an optimal (non-explainable) one, is $O(k^2)$.

After \cite{dasgupta2020explainable}, 
new algorithms, yielding to improved bounds on the
price of explainability, 
were proposed  \citep{laber2021price,DBLP:conf/icml/MakarychevS21,charikar2021near,esfandiari2021almost,gamlath2021nearly}.
The best known upper bound,
among those that only depend on $k$, is  $O(k \log k )$  from
\cite{esfandiari2021almost}. We note that this bound is 
nearly tight since the same paper also provides 
an $\Omega(k)$ lower bound. \citep{laber2022complexity} shows that the $k$-means explainable clustering problem is hard to approximate, thus consolidating the motivation for exploring heuristics for this problem. 

Empirical studies with algorithms for building
explainable partitions
can be found in 
\cite{frost2020exkmc,laber2021price}.
The former proposes the \textsc{ExKMC}
algorithm and  compares 
it  with \textsc{IMM}, \textsc{CART} \citep{breiman1984classification}, \textsc{KDTree} \citep{bentley1975multidimensional}, \textsc{CUBT} \citep{fraiman2013interpretable}, and \textsc{CLTree} \citep{liu2005clustering}. One  conclusion that
can be drawn from this study
is that  \textsc{IMM} and \textsc{ExKMC} outperform the other competitors
when the objective is building trees with exactly $k$ leaves. \textsc{ExKMC}, in contrast to  {\sc IMM}, is not limited to building trees with $k$ leaves, allowing partitions where the same cluster is associated with more than one leaf. This flexibility allows partitions with lower costs (though less explainable).  An algorithm with provable guarantees for this  scenario  was recently obtained in \cite{makarychev2021explainable}.

\cite{laber2021price} introduces a simple greedy algorithm,  \textsc{ExGreedy}, and shows that it produces partitions with lower costs than those produced by \textsc{IMM}.  We note that neither \cite{frost2020exkmc} nor \cite{laber2021price} analyze the produced trees in terms of their explainability. In  our experiments  we compare \textsc{IMM}, \textsc{ExGreedy}, and \textsc{ExKMC} against our method using different measures of explainability.

The aforementioned papers focus on the $k$-means clustering problem. However, a number of papers \citep{fraiman2013interpretable, bertsimas2018interpretable, saisubramanian2020balancing} propose  decision-tree algorithms to build partitions that optimize other measures.

The importance of building shallow trees for achieving interpretability has been previously discussed in \cite{blanco2020machine}, in which clustering and decision trees (constructed with the CART algorithm \citep{breiman1984classification}) are used to locally interpret the results of a black-box model.

\section{A strategy for building  shallow trees with low cost} \label{sec:algos-0}

Our strategy, denoted by {\sc ExShallow},  builds a   decision tree in a top-down fashion as shown in Algorithm \ref{alg:modo2}. 
As an input the strategy receives a set 
of points  ${\cal X}'$ and also a set  ${\cal S}'$ of $k$ representatives (denoted here by reference centers). 
We say that two cuts are equivalent with respect
to set ${\cal X}' \cup {\cal S}'$ if they are
associated with the same component (both are $i$-cuts for some $i$) and if they induce the same binary partition  on ${\cal X}' \cup {\cal S}'$. Note that there are at most $|{\cal X}' \cup {\cal S}'|d$ pairwise non-equivalent cuts. At each node the strategy evaluates 
\begin{equation} 
{\tt Price}(\gamma,{\cal X}',{\cal S}')
 + \lambda \cdot \depth(\gamma,{\cal X}',{\cal S}')
 \label{eq:greedycostfunction}
 \end{equation}
 for each cut $\gamma$ in the set 
of non-equivalent cuts  that separate at least
two reference centers from  ${\cal S}'$.
Then, it selects the cut $\gamma^*$ for which 
\eqref{eq:greedycostfunction} is minimum.

In Equation (\ref{eq:greedycostfunction}), {\tt Price}($\gamma$,${\cal X}'$,${\cal S}'$) and
  $\depth(\gamma$,${\cal X}'$,${\cal S}')$ (both detailed further below)
estimate how good $\gamma$ is for the goal of building a partition with low cost and 
with low values for $\waes / \wad$, respectively;
$\lambda$ is a trade-off parameter that we discuss
 in Section \ref{sec:trade-off-par}.
We note that $\depth$ stands for {\em Depth Explainability}. 

 
After selecting $\gamma^*$, the strategy  is recursively performed for each of the  groups of the binary partition induced by $\gamma^*$.
The recursion stops when ${\cal S}'$ contains only one reference center.
The initial set of reference centers
can be built by any 
algorithm for the (non-explainable) $k$-means
clustering problem, such as Lloyd's algorithm \citep{lloyd1982least}.

\begin{algorithm}[H]
  \caption{{\sc ExShallow}(${\cal X}'$, ${\cal S}'$)}
    ${\cal X}'$: set of points; ${\cal S}' $: set of reference centers
   \begin{algorithmic}[1]
  	
  	\small
	\IF{$|S'|=1$}
          	\STATE{Return ${\cal X}'$ and the single reference center in $S$'} \\
		\ELSE
    
    \STATE{ ${\cal C} \gets$ set of non-equivalent cuts w.r.t.
    ${\cal X}' \cup {\cal S}'$
    that separate at least two centers in ${\cal S}'$} 


            \STATE{
$\gamma^* \gets  \arg \min_{\gamma \in  {\cal C }}  \{ {\tt Price}(\gamma,{\cal X}', {\cal S}') + \lambda \cdot \depth(\gamma,{\cal X}', {\cal S}') \}$\label{line:GreedyChoice}
            }\\

            \STATE{$({\cal X}^*_L,{\cal X}^*_R) \gets $ partition of ${\cal X}'$  induced by $\gamma^*$} \\
            \STATE{$({\cal S}^*_L,{\cal S}^*_R) \gets $ partition of ${\cal S}'$  induced by $\gamma^*$} \\
    	    \STATE{Create a node $u$} \\
        	\STATE $u.{\tt LeftChild} \gets$ {\sc ExShallow}$({\cal X}^*_L, {\cal S}^*_L)$ \\
        	\STATE $u.{\tt RightChild} \gets$ {\sc ExShallow}$({\cal X}^*_R, {\cal S}^*_R)$ \\
            \STATE{Return the tree rooted at $u$} \\
        \ENDIF \\
  \end{algorithmic}
  \label{alg:modo2}
\end{algorithm}

Let ${\cal X'}$ and ${\cal S'}$ be, respectively, the sets of points and  centers that  reach some given node  in the decision tree. In addition,
let $\gamma$ be a cut that splits
 ${\cal X'}$ into groups ${\cal X'_L}$ and ${\cal X'_R}$ and splits  ${\cal S'}$ into groups ${\cal S'_L}$ and ${\cal S'_R}$,
each of them containing at least one reference center.
${\tt Price}(\gamma,{\cal X}', {\cal S}'$) is defined as 
\begin{equation}
\label{eq:price}
{\tt Price}(\gamma,{\cal X}', {\cal S}'):=
\frac{{\tt InducedCost}(\gamma,{\cal X}',{\cal S}')}{{\tt CurrentCost}({\cal X}',{\cal S}')},
\end{equation}
where
\begin{equation}
    \texttt{CurrentCost}({\cal X}',{\cal S}') :=   
     \sum_{\xvec \in {\cal X'}} \min_{\cvec \in  {\cal S'}} ||\xvec - \cvec||^2_2 
    \label{eq:cost_greedy0}
\end{equation}
and
\begin{equation}
\begin{split}
    \texttt{InducedCost}(\gamma,{\cal X}',{\cal S}'
) :=   
\left (
     \sum_{\xvec \in  {\cal X'_L}} \min_{\cvec \in  {\cal S'_L}} ||\xvec - \cvec||^2_2 + \sum_{\xvec \in  {\cal X'_R}} \min_{\cvec \in {\cal S'_R}} ||\xvec - \cvec||^2_2 
     \right );
    \label{eq:cost_greedy}
\end{split}
\end{equation}
that is, {\tt CurrentCost} and 
\texttt{InducedCost} 
give, respectively,  the cost of the partition before 
and after applying 
 cut $\gamma$. In both cases,
 each point is associated with the closest valid
 reference center.
We note that {\tt InducedCost} is the cost function used by the \textsc{ExGreedy} algorithm proposed in \cite{laber2021price}
to select a cut at each node.

To obtain   $\depth(\gamma,{\cal X}',{\cal S}'
)$, 
we first   calculate  
$\widehat{\wad}(\gamma,{\cal X}',{\cal S}'
) $,
an estimation of the quality of $\gamma$ for
finding a good tree in terms of $\wad$, and then we adjust $\widehat{\wad}(\gamma,{\cal X}',{\cal S}')$  to take into account the $\waes$.

Estimating whether a cut is good or not in terms of $\wad$ is a non-obvious task. For other metrics, such as the maximum depth of a tree, this is much simpler: the more balanced the cut, the better it is. To estimate the quality  of the cut $\gamma$ for our task, we efficiently compute \eqref{eq:WAD} for an auxiliary tree that is built specifically for this purpose.

More precisely,  $\widehat{\wad}(\gamma,{\cal X}',{\cal S}'
) $ is given by the return of the procedure presented in Algorithm \ref{alg:WAD}. $\texttt{EvalWAD}(N,K, r_p, r_c)$ returns the $\wad$ of a tree with $K$ leaves (corresponding to centers) for a set of $N$ points, where each node in the  tree splits the points and the centers in the same proportion as $\gamma$ does, that is, proportionally to $r_{p}=|{\cal X'}_L|/|{\cal X'}|$ and $r_{c}=|{\cal S'}_L|/|{\cal S'}|$, respectively. We note that these ratios do not change along the algorithm execution and that the resulting decision tree is just  a theoretical tree (which may not even be feasible for the instance under consideration), built to estimate how good the cut $\gamma$ is for the goal of minimizing Equation \eqref{eq:WAD}.
 
 \begin{algorithm}
  \caption{{\tt EvalWAD}(${N}$, $K$, $r_{p}$, $r_{c}$)}
   ${N}$: Current number of points; $K$: Current number of reference centers; $r_{p}$: Point-split ratio; $r_{c}$: Center-split ratio
   \begin{algorithmic}[1]
  	\small
	\IF{$ K=1$}
          	\STATE{Return 0} \\
		\ELSE
                        \STATE{$K_L \gets K \cdot r_{c}$  } 
                        \STATE{$K_R \gets K - K_L$}
                        \STATE{$N_L \gets N \cdot r_{p}$
                        }
                        \STATE{$N_R \gets N-N_L$}

            \STATE{Return $1+ (N_L \cdot {\tt EvalWAD}(N_L, K_L,r_p,r_c) +  N_R \cdot {\tt EvalWAD}(N_R, K_R,r_p,r_c) ) / N$} 
        \ENDIF \\
  \end{algorithmic}
  \label{alg:WAD}
\end{algorithm}
 
\begin{figure}
    \captionsetup[subfigure]{justification=centering}
     \centering
     \begin{subfigure}[b]{0.45\textwidth}
     \includegraphics[width=\textwidth]{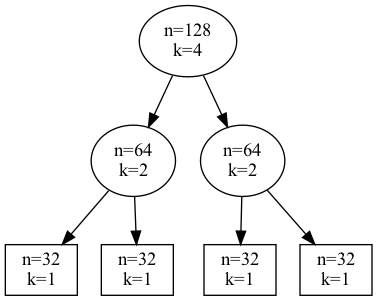}
     \subcaption{$r_{c} = r_{p} = 0.5$, $\wad = 2$}
     \label{fig:wad_balanced}
     \end{subfigure}
     \begin{subfigure}[b]{0.35\textwidth}
     \includegraphics[width=\textwidth]{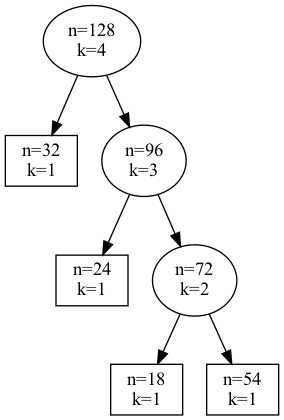}
     \subcaption{$r_{c} = r_{p} = 0.25$, $\wad = 2.31$}
     \label{fig:wad_unbalanced}
     \end{subfigure}
     \caption{Two trees generated by Algorithm \ref{alg:WAD} with different values of $r_{c}$ and $r_{p}$.}
     \label{fig:wad_compare}
 \end{figure}

 As an example, Figure \ref{fig:wad_compare} presents two such trees generated by Algorithm \ref{alg:WAD} for the same number of centers ($K=4$) and points ($N=128$), but different values of $r_{c}$ and $r_{p}$. In Figure \ref{fig:wad_balanced}, $r_{c} = r_{p} = 0.5$; as a result, the tree generated by Algorithm \ref{alg:WAD} has 4 leafs at level 2 with 25 points in each. In Figure \ref{fig:wad_unbalanced}, $r_{c} = r_{p} = 0.25$; as a result, the tree has 3 levels instead of 2, and most points are in one of the deepest leafs.  
 
The value of $\depth(\gamma,{\cal X'},{\cal S'})$
is given by the return of procedure
{\tt EvalDExp}($\gamma$, ${\cal X'}$,${\cal S'}$) 
 presented in Algorithm \ref{alg:Depth}.
To explain the procedure, let $v$ be the current node of the decision tree under construction.
Recall that if a cut $\gamma=(i,\theta)$ is applied on $v$ then it induces two
edges leaving $v$, one associated with condition
$x_i \leq \theta  $ and the other with condition
$x_i > \theta  $.
We say that an edge leaving $v$ is {\em killer}
if its associated condition turns
some non-redundant condition  
 in the path, from the root to $v$, into a redundant one. The procedure first determines
 which edges induced by $\gamma$ on $v$ are killer and,
 based on that, it adjusts
 the value of $\widehat{\wad}(\gamma,{\cal X'},{\cal S'})$ to take into
 account the metric $\waes$.
As an example, if only the left  edge leaving $v$
is killer then we discount $|{\cal X}'_L|/|{\cal X'}
 |$ from $\widehat{\wad}(\gamma,{\cal X'},{\cal S'})$ because one condition
 in the path from the root to $v$  becomes redundant
 to explain the clusters of the left
 subtree of $v$.

By design, ${\tt EvalDExp}$ prioritizes the choice of cuts at node $v$ that are associated with coordinates that have already been used by some cut in the path from the root to $v$. This way the strategy tends to produce redundant conditions and, therefore, to find good trees in terms of $\waes$.

\begin{algorithm}
  \caption{{\tt EvalDExp}($\gamma$, ${\cal X'}$, ${\cal S'}$) }
   $\gamma$: cut; ${\cal X'}$: set of points; ${\cal S'}$: set of centers 
   \begin{algorithmic}[1]
  	\small
    \STATE{$v \gets $ current node in the decision tree}
    \STATE{$({\cal X}'_L,{\cal X}'_R) \gets $ partition of ${\cal X}'$  induced by $\gamma$} \\
    \STATE{$({\cal S}'_L,{\cal S}'_R) \gets $ partition of ${\cal S}'$  induced by $\gamma$} \\
    \STATE{$r_p = {\cal X'}_L / {\cal X}'$}
    \STATE{$r_c = {\cal S'}_L / {\cal S}'$}
    \STATE{$\widehat{\wad} = {\tt EvalWAD}(|{\cal X'}|,|{\cal S'}|, r_p, r_c)$}
	\IF{no edge induced by $\gamma$ on  $v$ is killer}
	    \STATE{Return $\widehat{\wad}$} \\
    \ELSIF{only the left edge induced by $\gamma$ on $v$ is killer}
        \STATE{Return  $\widehat{\wad} - |{\cal X}'_L|/|{\cal X}'|$} \\
\ELSIF{only the right edge induced by $\gamma$ on $v$ is killer}
        \STATE{Return  $\widehat{\wad}- |{\cal X}'_R|/|{\cal X}'|$} \\
     \ELSE
        \STATE{Return  $\widehat{\wad} - 1$} \\
        \ENDIF \\
  \end{algorithmic}
  \label{alg:Depth}
\end{algorithm}

To summarize,  {\tt ExShallow}
follows the steps of  Algorithm \ref{alg:modo2}.
At line 5,
it calls  {\tt EvalDexp}, presented in Algorithm \ref{alg:Depth},
to evaluate $\depth(\gamma, {\cal X'},{\cal S'})$ and the value ${\tt Price}(\gamma, {\cal X'},{\cal S'})$
is calculated via Equations \ref{eq:price}, \ref{eq:cost_greedy0}
 and \ref{eq:cost_greedy}.

\subsection{Implementation details and time-complexity analysis for {\sc ExShallow}}
\label{sec:Implementation}
{\sc ExShallow} can be implemented in $O(n\cdot k \cdot d \cdot \wad({\cal D}))$ time, where $\wad({\cal D})$ is the $\wad$  of the decision tree ${\cal D}$  built by the algorithm. Given the set of points ${\cal X}$ and the reference centers ${\cal S}$, the algorithm first obtains $d$ sorted lists, where the $i$-th list corresponds to  the set of points in ${\cal X} \cup {\cal S} $ sorted by component $i$. This initial sorting step  takes $O(d (n+k) \log (n+k))$ time and it is only performed in the root of the tree.

Having the $d$ sorted lists at node $v$,  it is shown in  \cite{laber2021price} that ($\ref{eq:cost_greedy}$) can be computed for all valid cuts in ${\cal O}(d n_v  k_v)$ time,  where $n_v$ and $k_v$ are, respectively, the number of points and centers that reach $v$. In addition, the computation of $\widehat{\wad}$, via Algorithm \ref{alg:WAD}, takes $O(k_v)$ time per cut and, then, $O(d n_v  k_v)$ time for all cuts.

To find out which of the edges are killer in Algorithm \ref{alg:Depth},  we maintain a data structure, namely {\tt A}, with $2d$ entries. For each $i \in [d]$,  {\tt A}[$i$].left (resp. {\tt A}[i].right) stores the number of left (resp. right) edges that leave  $i$-nodes that lie in the path from the root to the current node. To determine if a left (resp. right) edge leaving an $i$-node is killer we test whether {\tt A}[$i$].left $>0$ (resp. {\tt A}[$i$].right $>0$ ) or not. In the positive case the edge is killer, otherwise it is not.
 
The data structure ${\tt A}$ can be updated in $O(1)$ time: if the chosen cut at node $v$ is an $i$-cut, then right before the recursive call at line 9 (resp. line 10)  of {\sc ExShallow} (Algorithm \ref{alg:modo2})  we increment by one unit  ${\tt A}[i]$.left (resp. ${\tt A}[i]$.right), and when we return from the recursion we decrease the respective counter by 1.

After selecting the cut at node $v$, the $d$ sorted lists for the children of $v$ are obtained in $O(n_v d)$ time from the sorted lists for $v$.

Thus, the total cost of the algorithm to build a tree ${\cal D}$  is  proportional to 
$$
\sum_{v \in {\cal D}} n_v \cdot d \cdot k_v \le \sum_{i=1}^n \ell_i \cdot d \cdot k,
$$ 
where $\ell_i$ is the depth of data point $i$ at ${\cal D}$. The rightmost term, however, is equal to  $\wad({\cal D}) \cdot n \cdot d \cdot k $.

The ${\cal O} (\wad({\cal D}) \cdot n \cdot d \cdot k)$  time complexity suggests that trees with low $\texttt{WAD}$ are faster to build -- which is good for our purposes, since by design our algorithm tries to build  trees with this property.

\subsection{Setting the trade-off parameter}
\label{sec:trade-off-par}

In a typical case, users are interested in obtaining an explainable clustering 
with low cost. To achieve this goal they have to properly set the value of $\lambda$.  One possibility is performing a brute-force 
search over some set of values  to find the
one that yields the most suitable tree. However, this could be computationally expensive and also non-practical from the users' perspective, 
as they would have to analyze many trees. 
Fortunately, as we explain, we can avoid that.

First we note that a  reasonable  interpretation for $\lambda$ is how much we are willing to (locally)
give up  of  cost, in percentage,  to reduce by one unit  the average size of the explanations. 
As an example, setting $\lambda=0.1$
means that we accept an additive loss of up to $10\%$ in terms of the partition cost to
have explanations one unit shorter on average.

Under this perspective, we shall avoid large values for $\lambda$,  since
partitions with high costs are not likely to produce
coherent clusters, and making incoherent explainable clusters would be useless. In fact, as we show in our experiments, by setting $\lambda$  to $0.03$ we obtain 
significant improvements over the existing methods.

A good property of   {\tt Price} (Equation \ref{eq:price}) is that
its value  for cuts of low {\tt InducedCost},
the most relevant ones,
lies in the interval [$1,4k+1$],
the same one in which both $\waes$ \eqref{eq:WAES} and $\wad$ \eqref{eq:WAD}  lie, except for a constant factor. 
Hence, we are trading off quantities with similar magnitudes, which is beneficial. This is formalized below.

\begin{lemma} \label{lem:cut} Let ${\cal X}'$ and ${\cal S}'$
 be the set of data points and reference
 centers that reach a given node $v$.
Then, there is a cut $\gamma'$ that satisfies
$1 \le {\tt Price}(\gamma',{\cal X}',{\cal S}' ) \le 4|{\cal S}'|+1$.
\end{lemma}
\begin{proof}
The lefthand side
follows because any 
assignment between points and reference centers
that is valid after applying a  cut is also valid before the cut, so that 
$\texttt{CurrentCost}({\cal X}',{\cal S}')
\le \texttt{InducedCost}(\gamma, {\cal X}',{\cal S}')$,
for every cut $\gamma$.

For the righthand side, we use the ideas from
\cite{dasgupta2020explainable}.  
Let $max_i$ and $min_i$ be the  maximum and minimum values of the $i$-th component among the centers in 
 ${\cal S}'$, respectively. Moreover, let $b_i=max_i -min_i$
and let $p(\gamma)$ be the number of points 
in ${\cal X}'$ that are separated
from their closest centers in ${\cal S}'$ 
when a cut $\gamma$ is employed.
We have that 
$$\texttt{InducedCost}(\gamma, {\cal X}',{\cal S}')
\le\texttt{CurrentCost}({\cal X}',{\cal S}') + p(\gamma) \sum_{i=1}^d b_i^2 .$$
The reason is that $\sum_{i=1}^d b_i^2$ is an upper bound on the contribution for the $k$-means cost of a point that is separated from its closest center.

On the other hand, it follows from 
Lemma 5.7 of \cite{dasgupta2020explainable}
that  
$$ \texttt{CurrentCost}({\cal X}',{\cal S}') \ge \frac{ p^*}{4|{\cal S}'|} \sum_{i=1}^d b_i^2, $$
where $p^*$ is the number of points separated
from their closest centers by the valid cut that separates the minimum number of points.

Thus, if $\gamma$ is a cut that
separates $p^*$ points from its closest centers, we
get that  $$ \texttt{InducedCost}(\gamma, {\cal X}',{\cal S}') \le (4|{\cal S'}|+1) \texttt{CurrentCost}({\cal X}',{\cal S}'), $$ establishing the result.
\end{proof}

\section{Experiments}
\label{sec:experiments}
In this section we report our experimental study.
We have two goals:
understanding the impact of $\lambda$ and,
most importantly, comparing our strategy  
with  other available proposals
for building explainable clustering \cite{dasgupta2020explainable, frost2020exkmc, laber2021price}.

These methods start with the reference centers 
of a partition for the unrestricted $k$-means clustering problem and, then, build a tree in a top-down fashion by selecting at each node a cut that separates at least two
reference centers. What distinguishes
them is the strategy employed to choose the cut: \textsc{IMM} \cite{dasgupta2020explainable}  selects the cut that minimizes the number of data points separated from their representatives; \textsc{ExKMC} \cite{frost2020exkmc} selects the cut that minimizes the overall $k$-means cost of the split when a single center (chosen from the original centers of the unrestricted solution) is assigned to all points in each side of the cut; and \textsc{ExGreedy} \cite{laber2021price}, as already mentioned, selects the cut that minimizes the {\tt InducedCost} given by Equation (\ref{eq:cost_greedy}).

In our evaluation, we considered 16  datasets of different sizes and characteristics, performing 10 or 30 seeded iterations in each of them, depending on the experiment. For each iteration, we find an unrestricted partition of the data by running Lloyd's algorithm \citep{lloyd1982least} with the ++ initialization \citep{arthur2006k}, as implemented in Python's \texttt{scikit-learn} package \cite{scikit-learn}. This unrestricted partition is provided to \textsc{IMM} and to \textsc{ExKMC}, as implemented in the \texttt{ExKMC} package \cite{frost2020exkmc}, and to \textsc{ExGreedy}, implemented as an extension of the \texttt{ExKMC} package and available in \url{https://github.com/lmurtinho/ExKMC} \cite{laber2021price}.
Then, we provide the same unrestricted partition to {\sc ExShallow}, available as {\tt ShallowTree} in the package of the same name available in \url{https://github.com/lmurtinho/ShallowTree}.

\subsection{Dataset summary}

Table \ref{tab:datasets} presents the size, dimension, and number of classes (which we use as the number of clusters) of the datasets in which we perform the experiments. All datasets are available online, and our code includes a script for retrieving and running tests on them. The number of instances, dimensions, and features is that of the final dataset used in our experiments (after removal of missing values and one-hot encoding of categorical variables, for instance). Most datasets are retrieved from OpenML \citep{OpenML2013} or UCI \citep{Dua2019UCI}. All datasets are anonimized and present no offensive content.

\begin{table*}
    \caption{Dataset summary: $n$ is the number of data points, $d$ is the dimension, and $k$ is the number of desired clusters.} 
    \vskip 0.15in

    \centering
    \begin{adjustbox}{width=1\textwidth}
    \begin{tabular}{l|r|r|r|c} 
        {\bf Dataset} & $n$ & $d$ & $k$ & {\bf Source}   \\
        \hline
        Anuran & 7,195  & 22 & 10 & UCI \\
        Avila & 20,867 & 10 & 12 & UCI \citep{de2018reliable} \\
        Beer & 1,514,999 & 5 & 104 & OpenML \\
        BNG (audiology) & 1,000,000 & 85 & 24 & OpenML \\
        Cifar10 & 60,000 & 3,072 & 10 & \cite{krizhevsky2009learning}  \\
        Collins & 1,000 & 19 & 30 & OpenML \\
        Covtype & 581,012 & 54 & 7 & OpenML \citep{collobert2002parallel}  \\
        Digits & 1797 & 64 & 10 & UCI \citep{alpaydin1998cascading} \\
        Iris & 150 & 4 & 3 & UCI \citep{fisher1936use} \\
        Letter & 20,000 & 16 & 26 & \cite{hsu2002comparison} \\
        Mice & 552 & 77 & 8 & OpenML \citep{higuera2015self} \\
        20Newsgroups & 18,846 & 1,069 & 20 &  \url{http://qwone.com/~jason/20Newsgroups/}\\
        Pendigits & 10,992 & 16 & 10 & UCI \\
        Poker & 1,025,010 & 10 & 10 & UCI \\
        Sensorless & 58,509 & 48 & 11 & UCI \\
        Vowel & 990 & 10 & 11 & UCI
    \end{tabular}
    \end{adjustbox}
    \label{tab:datasets}
\end{table*}

\subsection{Results}
\label{sec:results}

Table \ref{tab:full_results} shows the main results of our experiments for the 16 datasets and for 4 different explainable clustering algorithms: {\sc ExShallow}  with $\lambda = 0.03$,  \textsc{IMM} \citep{dasgupta2020explainable}, \textsc{ExKMC} \citep{frost2020exkmc}, and \textsc{ExGreedy} \citep{laber2021price}. For each dataset, we ran 30 seeded iterations of Lloyd's algorithm, and used the resulting (non explainable) partition as a starting point for each explainable clustering algorithm analyzed here. 

\afterpage{
    \clearpage
    \begin{landscape}
        \begin{table}[]
            \centering
            \caption{Full results for experiments for all datasets and algorithms. \textsc{ExShallow} is run with $\lambda = 0.03$. Best results for each dataset are in bold. For the partition cost, $\waes$, and $\wad$, values in red (blue) are statistically larger (smaller) than those of \textsc{ExShallow}, with a confidence level of 95\%. For the normalized information score, the results from the unexplained partition (via Lloyd's algorithm) are taken to be the ground truth, and values in red (blue) are statistically smaller (larger) than those of {\sc ExShallow}, with a confidence level of 95\%.}
            \vskip 0.15in
            \begin{adjustbox}{width=1.4\textwidth}
            \begin{tabular}{lr|rrrr|rrrr|rrrr|rrrr}
& & \multicolumn{4}{c|}{Normalized Partition Cost} & \multicolumn{4}{c|}{$\waes$} & \multicolumn{4}{c|}{$\wad$} & \multicolumn{4}{c}{Normalized Mutual Information} \\
\hline
{\bf Dataset}&$k$&ExShallow&ExGreedy&IMM&KMC&ExShallow&ExGreedy&IMM&KMC&ExShallow&ExGreedy&IMM&KMC&ExShallow&ExGreedy&IMM&KMC\\
Anuran&10&  \textcolor{black}{1.16}&{  \bf      \textcolor{blue}{1.15}}&  \textcolor{red}{1.28}&  \textcolor{red}{1.32}&  \textcolor{black}{3.75}&  \textcolor{red}{4.17}&  \textcolor{red}{5.37}&{  \bf      \textcolor{blue}{3.41}}&  \textcolor{black}{3.79}&  \textcolor{red}{4.27}&  \textcolor{red}{5.67}&{  \bf      \textcolor{blue}{3.41}}&  \textcolor{black}{0.70}&{  \bf      \textcolor{blue}{0.72}}&  \textcolor{black}{0.70}&  \textcolor{red}{0.64}\\
Avila&12&{  \bf      \textcolor{black}{1.05}}&  \textcolor{black}{1.05}&  \textcolor{red}{1.07}&  \textcolor{red}{1.18}&  \textcolor{black}{3.87}&  \textcolor{red}{5.58}&  \textcolor{red}{5.25}&{  \bf      \textcolor{blue}{3.26}}&  \textcolor{black}{4.60}&  \textcolor{red}{6.64}&  \textcolor{red}{6.61}&{  \bf      \textcolor{blue}{4.47}}&  \textcolor{black}{0.73}&  \textcolor{black}{0.72}&{  \bf      \textcolor{black}{0.73}}&  \textcolor{red}{0.68}\\
Beer&104&{  \bf      \textcolor{black}{1.16}}&  \textcolor{red}{1.19}&  \textcolor{red}{1.83}&  \textcolor{red}{1.27}&  \textcolor{black}{7.35}&  \textcolor{red}{8.13}&  \textcolor{red}{7.80}&{  \bf      \textcolor{blue}{6.34}}&  \textcolor{black}{10.47}&  \textcolor{red}{15.09}&  \textcolor{red}{54.31}&{  \bf      \textcolor{blue}{7.35}}&{  \bf      \textcolor{black}{0.83}}&  \textcolor{red}{0.82}&  \textcolor{red}{0.76}&  \textcolor{red}{0.81}\\
BNG&24&  \textcolor{black}{1.05}&{  \bf      \textcolor{blue}{1.02}}&  \textcolor{blue}{1.04}&  \textcolor{blue}{1.03}&{  \bf      \textcolor{black}{3.50}}&  \textcolor{red}{5.41}&  \textcolor{red}{8.82}&  \textcolor{red}{4.60}&{  \bf      \textcolor{black}{3.50}}&  \textcolor{red}{5.41}&  \textcolor{red}{11.82}&  \textcolor{red}{4.60}&  \textcolor{black}{0.29}&{  \bf      \textcolor{blue}{0.41}}&  \textcolor{red}{0.25}&  \textcolor{blue}{0.38}\\
Cifar10&10&{  \bf      \textcolor{black}{1.16}}&  \textcolor{red}{1.17}&  \textcolor{red}{1.22}&  \textcolor{red}{1.19}&{  \bf      \textcolor{black}{3.37}}&  \textcolor{red}{3.60}&  \textcolor{red}{5.70}&  \textcolor{red}{3.63}&{  \bf      \textcolor{black}{3.37}}&  \textcolor{red}{3.60}&  \textcolor{red}{5.70}&  \textcolor{red}{3.63}&{  \bf      \textcolor{black}{0.29}}&  \textcolor{red}{0.29}&  \textcolor{red}{0.25}&  \textcolor{red}{0.27}\\
Collins&30&  \textcolor{black}{1.18}&{  \bf      \textcolor{black}{1.17}}&  \textcolor{red}{1.23}&  \textcolor{red}{1.23}&{  \bf      \textcolor{black}{5.56}}&  \textcolor{red}{13.12}&  \textcolor{red}{12.81}&  \textcolor{black}{5.61}&  \textcolor{black}{5.86}&  \textcolor{red}{15.29}&  \textcolor{red}{17.00}&{  \bf      \textcolor{black}{5.83}}&{  \bf      \textcolor{black}{0.55}}&  \textcolor{red}{0.54}&  \textcolor{red}{0.54}&  \textcolor{red}{0.53}\\
Covtype&7&  \textcolor{black}{1.03}&  \textcolor{red}{1.03}&{  \bf      \textcolor{blue}{1.03}}&  \textcolor{red}{1.13}&  \textcolor{black}{2.61}&  \textcolor{red}{2.62}&  \textcolor{blue}{2.61}&{  \bf      \textcolor{blue}{2.45}}&  \textcolor{black}{3.15}&  \textcolor{red}{3.56}&  \textcolor{red}{3.55}&{  \bf      \textcolor{blue}{2.82}}&  \textcolor{black}{0.83}&  \textcolor{red}{0.83}&{  \bf      \textcolor{blue}{0.83}}&  \textcolor{red}{0.72}\\
Digits&10&{  \bf      \textcolor{black}{1.19}}&  \textcolor{red}{1.21}&  \textcolor{red}{1.23}&  \textcolor{red}{1.22}&  \textcolor{black}{3.96}&  \textcolor{red}{5.65}&  \textcolor{red}{5.36}&{  \bf      \textcolor{blue}{3.80}}&  \textcolor{black}{3.96}&  \textcolor{red}{5.65}&  \textcolor{red}{5.36}&{  \bf      \textcolor{blue}{3.80}}&{  \bf      \textcolor{black}{0.58}}&  \textcolor{red}{0.55}&  \textcolor{red}{0.55}&  \textcolor{red}{0.54}\\
Iris&3&{  \bf      \textcolor{black}{1.04}}&{  \bf      \textcolor{black}{1.04}}&{  \bf      \textcolor{black}{1.04}}&{  \bf      \textcolor{black}{1.04}}&  \textcolor{black}{1.67}&  \textcolor{black}{1.67}&{  \bf      \textcolor{blue}{1.44}}&{  \bf      \textcolor{blue}{1.44}}&{  \bf      \textcolor{black}{1.67}}&{  \bf      \textcolor{black}{1.67}}&{  \bf      \textcolor{black}{1.67}}&{  \bf      \textcolor{black}{1.67}}&{  \bf      \textcolor{black}{0.91}}&{  \bf      \textcolor{black}{0.91}}&{  \bf      \textcolor{black}{0.91}}&{  \bf      \textcolor{black}{0.91}}\\
Letter&26&{  \bf      \textcolor{black}{1.19}}&  \textcolor{red}{1.23}&  \textcolor{red}{1.30}&  \textcolor{red}{1.36}&{  \bf      \textcolor{black}{5.26}}&  \textcolor{red}{11.37}&  \textcolor{red}{12.64}&  \textcolor{red}{5.44}&{  \bf      \textcolor{black}{5.48}}&  \textcolor{red}{12.50}&  \textcolor{red}{14.85}&  \textcolor{black}{5.54}&{  \bf      \textcolor{black}{0.61}}&  \textcolor{red}{0.58}&  \textcolor{red}{0.56}&  \textcolor{red}{0.53}\\
Mice&8&{  \bf      \textcolor{black}{1.07}}&  \textcolor{red}{1.09}&  \textcolor{red}{1.12}&  \textcolor{red}{1.15}&  \textcolor{black}{3.17}&  \textcolor{red}{3.32}&  \textcolor{red}{3.53}&{  \bf      \textcolor{blue}{3.12}}&  \textcolor{black}{3.24}&  \textcolor{red}{3.58}&  \textcolor{red}{3.76}&{  \bf      \textcolor{blue}{3.13}}&{  \bf      \textcolor{black}{0.72}}&  \textcolor{black}{0.71}&  \textcolor{red}{0.71}&  \textcolor{red}{0.65}\\
20Newsgroups&20&  \textcolor{black}{1.05}&{  \bf      \textcolor{blue}{1.01}}&  \textcolor{blue}{1.01}&  \textcolor{blue}{1.01}&{  \bf      \textcolor{black}{1.12}}&  \textcolor{red}{15.61}&  \textcolor{red}{15.53}&  \textcolor{red}{13.80}&{  \bf      \textcolor{black}{1.22}}&  \textcolor{red}{15.63}&  \textcolor{red}{15.53}&  \textcolor{red}{13.80}&  \textcolor{black}{0.10}&  \textcolor{blue}{0.55}&{  \bf      \textcolor{blue}{0.56}}&  \textcolor{blue}{0.53}\\
Pendigits&10&{  \bf      \textcolor{black}{1.14}}&  \textcolor{red}{1.14}&  \textcolor{red}{1.24}&  \textcolor{red}{1.32}&  \textcolor{black}{3.70}&  \textcolor{red}{4.43}&  \textcolor{red}{4.31}&{  \bf      \textcolor{blue}{3.49}}&  \textcolor{black}{3.77}&  \textcolor{red}{4.46}&  \textcolor{red}{4.44}&{  \bf      \textcolor{blue}{3.50}}&  \textcolor{black}{0.77}&{  \bf      \textcolor{black}{0.77}}&  \textcolor{red}{0.72}&  \textcolor{red}{0.67}\\
Poker&10&{  \bf      \textcolor{black}{1.10}}&  \textcolor{red}{1.10}&  \textcolor{black}{1.10}&  \textcolor{red}{1.12}&  \textcolor{black}{3.35}&  \textcolor{red}{3.37}&  \textcolor{red}{3.37}&{  \bf      \textcolor{blue}{3.23}}&  \textcolor{black}{3.35}&  \textcolor{red}{3.37}&  \textcolor{red}{3.37}&{  \bf      \textcolor{blue}{3.23}}&  \textcolor{black}{0.41}&  \textcolor{black}{0.41}&{  \bf      \textcolor{blue}{0.41}}&  \textcolor{red}{0.40}\\
Sensorless&11&  \textcolor{black}{1.02}&{  \bf      \textcolor{black}{1.02}}&  \textcolor{red}{1.03}&  \textcolor{red}{1.07}&{  \bf      \textcolor{black}{2.99}}&  \textcolor{red}{4.24}&  \textcolor{red}{4.10}&  \textcolor{red}{3.99}&{  \bf      \textcolor{black}{3.84}}&  \textcolor{red}{4.52}&  \textcolor{red}{4.44}&  \textcolor{red}{4.07}&  \textcolor{black}{0.91}&  \textcolor{black}{0.92}&{  \bf      \textcolor{blue}{0.92}}&  \textcolor{red}{0.88}\\
Vowel&11&{  \bf      \textcolor{black}{1.21}}&  \textcolor{red}{1.25}&  \textcolor{red}{1.36}&  \textcolor{red}{1.29}&  \textcolor{black}{3.89}&  \textcolor{red}{5.26}&  \textcolor{red}{5.74}&{  \bf      \textcolor{blue}{3.63}}&  \textcolor{black}{3.94}&  \textcolor{red}{5.76}&  \textcolor{red}{6.41}&{  \bf      \textcolor{blue}{3.64}}&{  \bf      \textcolor{black}{0.58}}&  \textcolor{red}{0.55}&  \textcolor{red}{0.53}&  \textcolor{red}{0.52}\\
\hline
Median&&1.12&1.12&1.17&1.19&3.60&4.84&5.36&3.63&3.78&4.96&5.68&3.72&0.66&0.65&0.63&0.59
            \end{tabular}
            \end{adjustbox}
            \label{tab:full_results}
        \end{table}
    \end{landscape}
}

We also performed statistical tests (one-sided $t$-tests, assuming the same variance for both distributions, and with a confidence level of 95\%) to check the statistical significance of the difference between results from {\sc ExShallow}  and each of the other algorithms. Values in red (resp. blue) in Table \ref{tab:full_results} indicate that results for the algorithm in question are worse (resp. better) on average than those of {\sc ExShallow}  with a confidence level of 95\%.

The partition costs are normalized by the cost of the unrestricted partition used as a starting point for the explainable clustering algorithms.

In terms of average cost, {\sc ExShallow} beats (with 95$\%$ confidence) \textsc{ExGreedy} in 9 datasets, \textsc{IMM} in 11 and \textsc{KMC} in 13. It is beaten by at least one algorithm on 4 datasets, in two of them by less than $1\%$. Only for \texttt{BNG} and \texttt{20Newsgroups}  the partitions generated by {\sc ExShallow}  are clearly worse (by at most 4\%), and for both datasets {\sc ExShallow}  returns partitions that are much more explainable (in terms of $\wad$ and $\waes$) than those of the other algorithms.

In terms of $\waes$, {\sc ExShallow} outperforms \textsc{ExGreedy} and \textsc{IMM} on 15 and 14 datasets, respectively. For many datasets it is beaten by {\sc KMC}  by a small margin and, when this happens, it almost always beats {\sc KMC} in terms of partition cost, frequently by large margins. Observe the median of $\waes$ in the last line of Table \ref{tab:full_results}. Results for $\wad$ are similar, although {\sc KMC} more frequently outperforms {\sc ExShallow} in this metric.

We also report the normalized mutual information score (NMI) \citep{strehl2002cluster} of the partitions generated by the explainable algorithms, considering that the ground truth is the unrestrained partition from which they are derived; a value of 1 corresponds to a perfect correspondence between partitions. The partition generated by {\sc ExShallow} is the closest to the unrestrained one for 7 datasets, and it's as good as those generated by the other explainable algorithms in another one. {\sc ExShallow} returns the worst partition (in terms of NMI) for a single dataset, {\tt 20Newsgroups}.

In summary, our experiments suggest that {\sc ExShallow}  is almost always at least close to the best result in terms of both partition cost and explainability, and frequently has a significant advantage in at least one of these dimensions when compared to the other 3 algorithms (as can be seen in the results for {\tt Avila}, {\tt Collins}, {\tt Letter}, and {\tt Pendigits}, for instance).

To illustrate the last point made above, we refer back  to Figures \ref{fig:exg_tree} and \ref{fig:alg1_tree}. Both induce the same partition, but {\sc ExShallow} generates a more balanced tree with a smaller weighted depth. The maximum depth of the {\sc ExGreedy} tree is 7, against 4 for the {\sc ExShallow}  tree. The largest cluster in the partition (cluster 10, with 5,605 elements) has depth 7 in the {\sc ExGreedy} tree, and explanation size 6 (condition $EX > 0.440$ at depth 3 is made redundant by condition $EX > 0.574$ at depth 4); in the {\sc ExShallow} tree, it has both depth and explanation size 4.

\subsection{Sensitivity of cost and weighted depth to variations in $\lambda$}

\begin{figure}[t]
        \includegraphics[width=\textwidth]{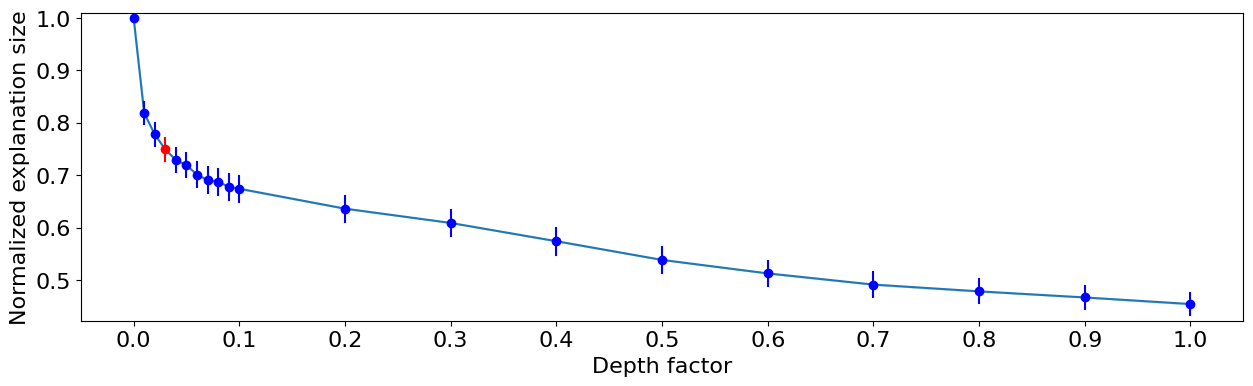}
        \caption{Mean $\waes$ per depth factor (for all datasets), normalized by the results for $\lambda = 0$ for each dataset. Error bars (with a confidence interval of 95\%) are calculated using Python's \texttt{scipy} package \citep{2020SciPy-NMeth}.}
        \label{fig:norm_exp_size}
\end{figure}

Figure \ref{fig:norm_exp_size} shows how the average $\waes$ of the partitions produced by {\sc ExShallow}  changes as $\lambda$ increases. To allow for a comparison between datasets, the values are normalized by those of the tree when $\lambda = 0$ (i.e., when depth is not taken into account by our cost function). For each dataset, we ran 10 seeded iterations of Lloyd's algorithm and used the resulting  partitions as a starting point for each instance of {\sc ExShallow}  with different values of $\lambda$.

{\sc ExShallow} behaves as expected, with larger values of $\lambda$  associated with  trees
having lower $\waes$, on average. (Results for $\wad$ are omitted as they are very similar in terms of correlation with $\lambda$.) We observe a sharp drop for small increments of $\lambda$ when starting from zero. The  red value is 0.03, the one employed in the previous experiments.

\begin{figure}[t]
        \includegraphics[width=\textwidth]{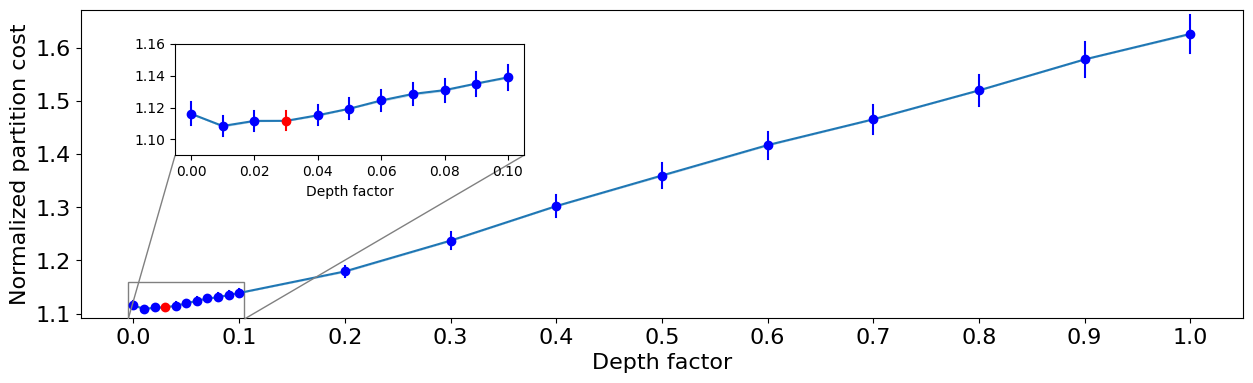}
        \caption{Mean normalized partition cost per depth factor (for all datasets), normalized by the cost of the unrestricted partition used to build the explainable partition. Error bars (with a confidence interval of 95\%) are calculated using \texttt{scipy}.}
        \label{fig:mean_norm_cost}
\end{figure}

Figure \ref{fig:mean_norm_cost} shows how the mean cost of the partitions produced by our algorithm changes as $\lambda$ increases. To allow for a comparison between datasets, the costs are normalized by the cost of the unrestricted partition generated by Lloyd's algorithm. The behavior is, in general, the expected one, with larger values of $\lambda$  associated with higher costs. 

Combining these figures leads to the important, and perhaps surprising,
empirical conclusion that working with a small $\lambda$ is very beneficial, as it significantly reduces the average weighted depth and explanation size without increasing the average cost of the partition.

\subsection{Calibrating the trade-off between partition quality and explainability}

The results presented in Figures \ref{fig:norm_exp_size} and \ref{fig:mean_norm_cost} suggest that calibrating $\lambda$ may lead to significant improvements when {\sc ExShallow} does not initially return partitions that are satisfactory in terms of either quality (cost) or explainability ($\waes$ and/or $\wad$). We believe the results presented in Table \ref{tab:full_results} indicate {\sc ExShallow} ``out of the box'' is at least competitive with, and arguably superior to, the most recent comparable algorithms presented and evaluated in the literature, but there is some room for improvement. For instance, although the partition induced by {\sc ExShallow} for the {\tt 20Newsgroups} dataset is much more explainable than those induced by the competition, the quality of the partition (both in terms of cost and NMI) suffers from it; and in many cases {\sc KMC} induces partitions that are slightly more explainable, although their quality tends to be worse.

We can use the $\lambda$ parameter to adjust the trade-off between partition quality and explainability in {\sc ExShallow}, something that is not possible in the other algorithms presented here. To do so, we devised a simple binary search strategy, that starts from our default value of $\lambda = 0.03$ and then, if necessary, decreases it to try and find a partition with smaller cost, or increases it to try and find a partition with smaller $\waes$. Given a goal cost $c^*$ and a goal $\waes$ $w^*$, the binary search aims to find a partition with cost $c \leq c^*$ and $\waes$ $w \leq w^*$; if it is unable to do so, it returns the partition with the smallest $\waes$ given that its cost does not exceed $c^*$.

We present the results of this binary search, over 30 seeded iterations for each algorithm, in Table \ref{tab:opt_kmc}. Considering that {\sc KMC} frequently beats {\sc ExShallow} in terms of $\waes$, we used its results as our goal; the idea being to check if we can ``dominate'' its results (i.e., induce partitions that have, on average, both smaller costs and explanation sizes) in the datasets under analysis. 

In terms of cost, {\sc ExShallow$^*$} ({\sc ExShallow} with $\lambda$ optimized by the procedure described above) beats {\sc KMC} in all but two datasets, where both algorithms are tied; in terms of $\waes$, {\sc ExShallow$^*$} beats {\sc KMC} in 13 datasets and is beaten by it in 3. Most notably, in the two datasets ({\tt BNG} and {\tt 20Newsgroups}) for which {\sc KMC} induces less costly partitions than {\sc ExShallow}, {\sc ExShallow$^*$} induces partitions that beat the ones generated by {\sc KMC} in both dimensions.

\begin{table}[]
    \centering
    \caption{Comparison between results for {\sc KMC} and {\sc ExShallow$^*$} ({\sc ExShallow} with $\lambda$ optimized via binary search to find a better partition than {\sc KMC}'s in terms of both cost and $\waes$). Best results for each dataset are in bold. For the normalized partition cost (NPC), $\waes$, and $\wad$, values in red (blue) are statistically larger (smaller) than those of \textsc{ExShallow$^*$}, with a confidence level of 95\%. For the normalized information score (NMI), the results from the unexplained partition (via Lloyd's algorithm) are taken to be the ground truth, and values in red (blue) are statistically smaller (larger) than those of {\sc ExShallow$^*$}, with a confidence level of 95\%.}
    \vskip 0.15in
    \begin{adjustbox}{width=1\textwidth}
    \begin{tabular}{lr|rr|rr|rr|rr}
& & \multicolumn{2}{c|}{NPC} & \multicolumn{2}{c|}{$\waes$} & \multicolumn{2}{c|}{$\wad$} & \multicolumn{2}{c}{NMI} \\
\hline
{\bf Dataset}&$k$&{\sc KMC}&{\sc ExShallow$^*$}&{\sc KMC}&{\sc ExShallow$^*$}&{\sc KMC}&{\sc ExShallow$^*$}&{\sc KMC}&{\sc ExShallow$^*$}\\
Anuran&10&  \textcolor{red}{1.32}&  \textcolor{black}{\bf 1.20}&  \textcolor{red}{3.41}&{  \bf      \textcolor{black}{\bf 3.19}}&  \textcolor{red}{3.41}&{  \bf      \textcolor{black}{3.33}}&  \textcolor{red}{0.64}&  \textcolor{black}{\bf 0.68}\\
Avila&12&  \textcolor{red}{1.18}&  \textcolor{black}{\bf 1.15}&  \textcolor{black}{3.26}&{  \bf      \textcolor{black}{3.24}}&  \textcolor{black}{4.47}&{  \bf      \textcolor{black}{3.76}}&  \textcolor{blue}{\bf 0.68}&  \textcolor{black}{0.64}\\
Beer&104&  \textcolor{red}{1.27}&  \textcolor{black}{\bf 1.22}&{  \bf      \textcolor{blue}{6.34}}&  \textcolor{black}{7.25}&{  \bf      \textcolor{blue}{7.35}}&  \textcolor{black}{10.53}&  \textcolor{red}{0.81}&  \textcolor{black}{\bf 0.82}\\
BNG&24&  \textcolor{red}{1.03}&{  \bf      \textcolor{black}{1.02}}&  \textcolor{red}{4.60}&  \textcolor{black}{\bf 4.50}&  \textcolor{red}{4.60}&  \textcolor{black}{\bf 4.50}&  \textcolor{black}{0.38}&{       \textcolor{black}{0.38}}\\
Cifar10&10&  \textcolor{red}{1.19}&{  \bf      \textcolor{black}{1.16}}&  \textcolor{red}{3.63}&{  \bf      \textcolor{black}{3.37}}&  \textcolor{red}{3.63}&{  \bf      \textcolor{black}{3.37}}&  \textcolor{red}{0.27}&{  \bf      \textcolor{black}{0.29}}\\
Collins&30&  \textcolor{red}{1.23}&  \textcolor{black}{\bf 1.20}&  \textcolor{red}{5.61}&{  \bf      \textcolor{black}{4.97}}&  \textcolor{red}{5.83}&{  \bf      \textcolor{black}{5.42}}&  {0.53}&  \textcolor{black}{ 0.53}\\
Covtype&7&  \textcolor{red}{1.13}&  \textcolor{black}{\bf 1.12}&  \textcolor{red}{2.45}&{  \bf      \textcolor{black}{2.44}}&  \textcolor{red}{2.82}&{  \bf      \textcolor{black}{2.65}}&  \textcolor{red}{0.72}&  \textcolor{black}{\bf 0.75}\\
Digits&10&  \textcolor{red}{1.22}&  \textcolor{black}{\bf 1.19}&  \textcolor{red}{3.80}&{  \bf      \textcolor{black}{3.65}}&  \textcolor{red}{3.80}&{  \bf      \textcolor{black}{3.65}}&  \textcolor{red}{0.54}&  \textcolor{black}{\bf 0.56}\\
Iris&3&{  \textcolor{black}{1.04}}&{      \textcolor{black}{1.04}}&{  \bf      \textcolor{blue}{1.44}}&  \textcolor{black}{1.67}&{        {1.67}}&{        \textcolor{black}{1.67}}&{        \textcolor{black}{0.91}}&{        \textcolor{black}{0.91}}\\
Letter&26&  \textcolor{red}{1.36}&  \textcolor{black}{\bf 1.24}&  \textcolor{red}{5.44}&{  \bf      \textcolor{black}{4.81}}&  \textcolor{red}{5.54}&{  \bf      \textcolor{black}{5.02}}&  \textcolor{red}{0.53}&  \textcolor{black}{\bf 0.58}\\
Mice&8&  \textcolor{red}{1.15}&  \textcolor{black}{\bf 1.10}&  \textcolor{red}{3.12}&{  \bf      \textcolor{black}{2.97}}&  \textcolor{red}{3.13}&{  \bf      \textcolor{black}{3.11}}&  \textcolor{red}{0.65}&  \textcolor{black}{\bf 0.70}\\
20Newsgroups&20&  {1.01}&{ 1.01}&  \textcolor{red}{13.80}&  \textcolor{black}{\bf 13.45}&  \textcolor{red}{13.80}&  \textcolor{black}{\bf 13.78}&{ \textcolor{black}{0.53}}&  \textcolor{black}{0.53}\\
Pendigits&10&  \textcolor{red}{1.32}&  \textcolor{black}{\bf 1.15}&  \textcolor{red}{3.49}&{  \bf      \textcolor{black}{\bf 3.28}}&  \textcolor{red}{3.50}&{  \bf      \textcolor{black}{3.37}}&  \textcolor{red}{0.67}&  \textcolor{black}{\bf 0.75}\\
Poker&10&  \textcolor{red}{1.12}&  \textcolor{black}{\bf 1.11}&{  \bf      \textcolor{blue}{3.23}}&  \textcolor{black}{3.33}&{  \bf      \textcolor{blue}{3.23}}&  \textcolor{black}{3.33}&  {0.40}&  {0.40}\\
Sensorless&11&  \textcolor{red}{1.07}&{  \bf      \textcolor{black}{\bf 1.02}}&  \textcolor{red}{3.99}&{  \bf      \textcolor{black}{\bf 2.99}}&  \textcolor{red}{4.07}&{  \bf      \textcolor{black}{\bf 3.84}}&  \textcolor{red}{0.88}&{  \bf      \textcolor{black}{0.91}}\\
Vowel&11&  \textcolor{red}{1.29}&  \textcolor{black}{\bf 1.24}&  \textcolor{red}{3.63}&{  \bf      \textcolor{black}{\bf 3.41}}&  \textcolor{red}{3.64}&{  \bf      \textcolor{black}{3.50}}&  \textcolor{red}{0.52}&  \textcolor{black}{\bf 0.56}\\
\hline
Median&&1.17&1.15&3.63&3.35&3.72&3.58&0.59&0.61\\
    \end{tabular}
    \end{adjustbox}
    \label{tab:opt_kmc}
\end{table}

\subsection{Running times}

Table \ref{tab:run_times} presents the average running times, over 30 seeded iterations, for each dataset and algorithm -- including Lloyd's algorithm ({\sc KMeans}), which finds the  partition used as a starting point for all four algorithms, and {\sc ExShallow$^*$}. For all explainable algorithms, we add to their running time that of {\sc KMeans}, as an unrestrained partitioned is needed as a starting point for them to find an explainable partition. Disregarding {\sc ExShallow$^*$}, which can be  expensive (as it performs several iterations of {\sc ExShallow}),  {\sc ExGreedy} is the slowest explainable algorithm for all datasets except {\tt Beer}, for which both {\sc IMM} and {\sc KMC} are slower. {\sc ExShallow}'s running times are typically closer to those of {\sc ExGreedy} than those of {\sc IMM} and {\sc KMC}, which tend to be faster. Overall, we do not perceive running time to be a significant hindrance in choosing {\sc ExShallow}  over the other explainable clustering algorithms analyzed here, particularly due to the overhead imposed by initially running {\sc KMeans}.

\begin{table}
\centering
\caption{Average running times (in seconds) for each algorithm and dataset, including Lloyd's algorithm ({\sc kmeans}).  Experiments were performed on 8 484 Intel Core i7-4790 processors @3.60GHz with 32 GB of RAM, running Ubuntu 20.04.3 LTS.}

\vskip 0.15in

\begin{adjustbox}{width=1\textwidth}
\begin{tabular}{lr|rrrrrr}
\hline
 {\bf Dataset}&$k$&{\sc K-means}&{\sc ExShallow}&{\sc ExShallow$^*$}&{\sc ExGreedy} &{\sc IMM} &{\sc KMC}\\
Anuran&10&0.46&0.75&0.74&0.74&0.56&0.62\\
Avila&12&1.44&1.84&3.81&1.93&1.60&1.77\\
Beer&104&722.65&730.90&938.72&731.75&752.97&760.48\\
BNG &24&896.97&1033.53&1669.02&1068.84&930.79&956.54\\
Cifar10&10&348.71&550.68&550.83&562.80&416.71&437.85\\
Collins&30&0.34&0.46&0.50&0.52&0.38&0.41\\
Covtype&7&36.88&59.30&58.01&61.63&42.51&48.76\\
Digits&10&0.27&0.44&0.44&0.48&0.32&0.36\\
Iris&3&0.02&0.02&0.06&0.02&0.02&0.02\\
Letter&26&4.15&4.83&4.84&5.55&4.40&4.57\\
Mice&8&0.14&0.24&0.41&0.24&0.16&0.18\\
20Newsgroups&20&47.38&57.51&311.83&107.11&54.56&63.59\\
Pendigits&10&0.74&1.01&1.02&1.04&0.83&0.92\\
Poker&10&83.67&97.32&328.40&97.30&87.32&95.67\\
Sensorless&11&3.11&7.21&7.22&7.69&4.04&5.00\\
Vowel&11&0.18&0.20&0.24&0.21&0.19&0.19\\
\hline
Median&10.50&2.28&2.82&3.81&3.22&2.48&2.65\\
\end{tabular}
\end{adjustbox}
\label{tab:run_times}
\end{table}

\section{Conclusions}

We discussed how explainable an ``explainable partition'' actually is, by analyzing the average depth of its underlying decision tree and the average number of rules needed to explain each cluster in the partition (both metrics being weighted by the number of points assigned to each leaf/cluster). In most of the previous work on explainable clustering via decision trees,  measures related to the depths of the leaves were largely ignored.

We present {\sc ExShallow}, a  simple and efficient  algorithm that seeks to minimize both the cost of the resulting explainable partition  and 
the aforementioned metrics.  
The algorithm has a tunable  parameter $\lambda$ that
allows to trade-off cost and explainability.
Our experiments suggest that by
working with a (fixed) small value for $\lambda$, {\sc ExShallow}   produces partitions that are at least as good as, and many times significantly better than, those obtained by 
the available methods in the literature.
Thus, we understand that it is a valuable tool for those interested in explainable partitions that optimize the quite popular $k$-means cost.

\begin{singlespace}
 \bibliographystyle{elsarticle-num} 
 \bibliography{cas-refs}

\begin{thebibliography}{10}
\expandafter\ifx\csname url\endcsname\relax
  \def\url#1{\texttt{#1}}\fi
\expandafter\ifx\csname urlprefix\endcsname\relax\def\urlprefix{URL }\fi
\expandafter\ifx\csname href\endcsname\relax
  \def\href#1#2{#2} \def\path#1{#1}\fi

\bibitem{DBLP:journals/jair/BurkartH21}
N.~Burkart, M.~F. Huber, \href{https://doi.org/10.1613/jair.1.12228}{A survey
  on the explainability of supervised machine learning}, J. Artif. Intell. Res.
  70 (2021) 245--317.
\newblock \href {https://doi.org/10.1613/jair.1.12228}
  {\path{doi:10.1613/jair.1.12228}}.
\newline\urlprefix\url{https://doi.org/10.1613/jair.1.12228}

\bibitem{DBLP:journals/eswa/PiltaverLGM16}
R.~Piltaver, M.~Lustrek, M.~Gams, S.~Martincic{-}Ipsic,
  \href{https://doi.org/10.1016/j.eswa.2016.06.009}{What makes classification
  trees comprehensible?}, Expert Syst. Appl. 62 (2016) 333--346.
\newblock \href {https://doi.org/10.1016/j.eswa.2016.06.009}
  {\path{doi:10.1016/j.eswa.2016.06.009}}.
\newline\urlprefix\url{https://doi.org/10.1016/j.eswa.2016.06.009}

\bibitem{de2018reliable}
C.~De Stefano, M.~Maniaci, F.~Fontanella, A.~Scotto di Freca,
  \href{https://www.sciencedirect.com/science/article/pii/S0952197618300721}{Reliable
  writer identification in medieval manuscripts through page layout features:
  The “{A}vila” {B}ible case}, Engineering Applications of Artificial
  Intelligence 72 (2018) 99--110.
\newblock \href
  {https://doi.org/https://doi.org/10.1016/j.engappai.2018.03.023}
  {\path{doi:https://doi.org/10.1016/j.engappai.2018.03.023}}.
\newline\urlprefix\url{https://www.sciencedirect.com/science/article/pii/S0952197618300721}

\bibitem{laber2021price}
E.~S. Laber, L.~Murtinho,
  \href{https://proceedings.mlr.press/v139/laber21a.html}{On the price of
  explainability for some clustering problems}, in: M.~Meila, T.~Zhang (Eds.),
  Proceedings of the 38th International Conference on Machine Learning, Vol.
  139 of Proceedings of Machine Learning Research, PMLR, 2021, pp. 5915--5925.
\newline\urlprefix\url{https://proceedings.mlr.press/v139/laber21a.html}

\bibitem{dasgupta2020explainable}
M.~Moshkovitz, S.~Dasgupta, C.~Rashtchian, N.~Frost,
  \href{https://proceedings.mlr.press/v119/moshkovitz20a.html}{Explainable
  $k$-means and $k$-medians clustering}, in: H.~D. III, A.~Singh (Eds.),
  Proceedings of the 37th International Conference on Machine Learning, Vol.
  119 of Proceedings of Machine Learning Research, PMLR, 2020, pp. 7055--7065.
\newline\urlprefix\url{https://proceedings.mlr.press/v119/moshkovitz20a.html}

\bibitem{frost2020exkmc}
N.~Frost, M.~Moshkovitz, C.~Rashtchian,
  \href{https://arxiv.org/abs/2006.02399}{Ex{KMC}: Expanding explainable
  $k$-means clustering}, arXiv (2020).
\newblock \href {https://doi.org/10.48550/ARXIV.2006.02399}
  {\path{doi:10.48550/ARXIV.2006.02399}}.
\newline\urlprefix\url{https://arxiv.org/abs/2006.02399}

\bibitem{DBLP:conf/icml/MakarychevS21}
K.~Makarychev, L.~Shan,
  \href{http://proceedings.mlr.press/v139/makarychev21a.html}{Near-optimal
  algorithms for explainable k-medians and k-means}, in: M.~Meila, T.~Zhang
  (Eds.), Proceedings of the 38th International Conference on Machine Learning,
  {ICML} 2021, 18-24 July 2021, Virtual Event, Vol. 139 of Proceedings of
  Machine Learning Research, {PMLR}, 2021, pp. 7358--7367.
\newline\urlprefix\url{http://proceedings.mlr.press/v139/makarychev21a.html}

\bibitem{charikar2021near}
M.~Charikar, L.~Hu,
  \href{https://epubs.siam.org/doi/abs/10.1137/1.9781611977073.101}{Near-optimal
  explainable $k$-means for all dimensions}, in: Proceedings of the 2022 Annual
  ACM-SIAM Symposium on Discrete Algorithms (SODA), 2022, pp. 2580--2606.
\newblock \href
  {http://arxiv.org/abs/https://epubs.siam.org/doi/pdf/10.1137/1.9781611977073.101}
  {\path{arXiv:https://epubs.siam.org/doi/pdf/10.1137/1.9781611977073.101}},
  \href {https://doi.org/10.1137/1.9781611977073.101}
  {\path{doi:10.1137/1.9781611977073.101}}.
\newline\urlprefix\url{https://epubs.siam.org/doi/abs/10.1137/1.9781611977073.101}

\bibitem{esfandiari2021almost}
H.~Esfandiari, V.~Mirrokni, S.~Narayanan,
  \href{https://epubs.siam.org/doi/abs/10.1137/1.9781611977073.103}{Almost
  tight approximation algorithms for explainable clustering}, in: Proceedings
  of the 2022 Annual ACM-SIAM Symposium on Discrete Algorithms (SODA), 2022,
  pp. 2641--2663.
\newblock \href
  {http://arxiv.org/abs/https://epubs.siam.org/doi/pdf/10.1137/1.9781611977073.103}
  {\path{arXiv:https://epubs.siam.org/doi/pdf/10.1137/1.9781611977073.103}},
  \href {https://doi.org/10.1137/1.9781611977073.103}
  {\path{doi:10.1137/1.9781611977073.103}}.
\newline\urlprefix\url{https://epubs.siam.org/doi/abs/10.1137/1.9781611977073.103}

\bibitem{gamlath2021nearly}
B.~Gamlath, X.~Jia, A.~Polak, O.~Svensson,
  \href{https://proceedings.neurips.cc/paper/2021/file/f24ad6f72d6cc4cb51464f2b29ab69d3-Paper.pdf}{Nearly-tight
  and oblivious algorithms for explainable clustering}, in: M.~Ranzato,
  A.~Beygelzimer, Y.~Dauphin, P.~Liang, J.~W. Vaughan (Eds.), Advances in
  Neural Information Processing Systems, Vol.~34, Curran Associates, Inc.,
  2021, pp. 28929--28939.
\newline\urlprefix\url{https://proceedings.neurips.cc/paper/2021/file/f24ad6f72d6cc4cb51464f2b29ab69d3-Paper.pdf}

\bibitem{laber2022complexity}
E.~S. Laber, \href{https://arxiv.org/abs/2208.09643}{The computational
  complexity of some explainable clustering problems} (2022).
\newblock \href {https://doi.org/10.48550/ARXIV.2208.09643}
  {\path{doi:10.48550/ARXIV.2208.09643}}.
\newline\urlprefix\url{https://arxiv.org/abs/2208.09643}

\bibitem{breiman1984classification}
L.~Breiman, J.~H. Friedman, R.~A. Olshen, C.~J. Stone,
  \href{https://doi.org/10.1201/9781315139470}{Classification and Regression
  Trees}, Wadsworth and Brooks, Monterey, CA, 1984.
\newline\urlprefix\url{https://doi.org/10.1201/9781315139470}

\bibitem{bentley1975multidimensional}
J.~L. Bentley, \href{https://doi.org/10.1145/361002.361007}{Multidimensional
  binary search trees used for associative searching}, Commun. ACM 18~(9)
  (1975) 509–517.
\newblock \href {https://doi.org/10.1145/361002.361007}
  {\path{doi:10.1145/361002.361007}}.
\newline\urlprefix\url{https://doi.org/10.1145/361002.361007}

\bibitem{fraiman2013interpretable}
R.~Fraiman, B.~Ghattas, M.~Svarc,
  \href{https://doi.org/10.1007/s11634-013-0129-3}{Interpretable clustering
  using unsupervised binary trees}, Adv. Data Anal. Classif. 7~(2) (2013)
  125--145.
\newblock \href {https://doi.org/10.1007/s11634-013-0129-3}
  {\path{doi:10.1007/s11634-013-0129-3}}.
\newline\urlprefix\url{https://doi.org/10.1007/s11634-013-0129-3}

\bibitem{liu2005clustering}
B.~Liu, Y.~Xia, P.~Yu, \href{https://doi.org/10.1007/11362197_5}{Clustering via
  decision tree construction}, in: W.~Chu, T.~Young~Lin (Eds.), Foundations and
  Advances in Data Mining, Springer Berlin Heidelberg, Berlin, Heidelberg,
  2005, pp. 97--124.
\newblock \href {https://doi.org/10.1007/11362197_5}
  {\path{doi:10.1007/11362197_5}}.
\newline\urlprefix\url{https://doi.org/10.1007/11362197_5}

\bibitem{makarychev2021explainable}
K.~Makarychev, L.~Shan,
  \href{https://doi.org/10.1145/3519935.3520056}{Explainable k-means: Don’t
  be greedy, plant bigger trees!}, in: Proceedings of the 54th Annual ACM
  SIGACT Symposium on Theory of Computing, STOC 2022, Association for Computing
  Machinery, New York, NY, USA, 2022, p. 1629–1642.
\newblock \href {https://doi.org/10.1145/3519935.3520056}
  {\path{doi:10.1145/3519935.3520056}}.
\newline\urlprefix\url{https://doi.org/10.1145/3519935.3520056}

\bibitem{bertsimas2018interpretable}
D.~Bertsimas, A.~Orfanoudaki, H.~Wiberg,
  \href{https://arxiv.org/abs/1812.00539}{Interpretable clustering via optimal
  trees}, arXiv (2018).
\newblock \href {https://doi.org/10.48550/ARXIV.1812.00539}
  {\path{doi:10.48550/ARXIV.1812.00539}}.
\newline\urlprefix\url{https://arxiv.org/abs/1812.00539}

\bibitem{saisubramanian2020balancing}
S.~Saisubramanian, S.~Galhotra, S.~Zilberstein,
  \href{https://doi.org/10.1145/3375627.3375843}{Balancing the tradeoff between
  clustering value and interpretability}, in: Proceedings of the AAAI/ACM
  Conference on AI, Ethics, and Society, Association for Computing Machinery,
  New York, NY, USA, 2020, p. 351–357.
\newline\urlprefix\url{https://doi.org/10.1145/3375627.3375843}

\bibitem{blanco2020machine}
A.~Blanco-Justicia, J.~Domingo-Ferrer, S.~Martínez, D.~Sánchez,
  \href{https://www.sciencedirect.com/science/article/pii/S0950705120300368}{Machine
  learning explainability via microaggregation and shallow decision trees},
  Knowledge-Based Systems 194 (2020) 105532.
\newblock \href {https://doi.org/https://doi.org/10.1016/j.knosys.2020.105532}
  {\path{doi:https://doi.org/10.1016/j.knosys.2020.105532}}.
\newline\urlprefix\url{https://www.sciencedirect.com/science/article/pii/S0950705120300368}

\bibitem{lloyd1982least}
S.~Lloyd, \href{https://doi.org/10.1109/TIT.1982.1056489}{Least squares
  quantization in {PCM}}, IEEE Transactions on Information Theory 28~(2) (1982)
  129--137.
\newblock \href {https://doi.org/10.1109/TIT.1982.1056489}
  {\path{doi:10.1109/TIT.1982.1056489}}.
\newline\urlprefix\url{https://doi.org/10.1109/TIT.1982.1056489}

\bibitem{arthur2006k}
D.~Arthur, S.~Vassilvitskii,
  \href{https://dl.acm.org/doi/10.5555/1283383.1283494}{K-means++: The
  advantages of careful seeding}, in: Proceedings of the Eighteenth Annual
  ACM-SIAM Symposium on Discrete Algorithms, SODA '07, Society for Industrial
  and Applied Mathematics, USA, 2007, p. 1027–1035.
\newline\urlprefix\url{https://dl.acm.org/doi/10.5555/1283383.1283494}

\bibitem{scikit-learn}
F.~Pedregosa, G.~Varoquaux, A.~Gramfort, V.~Michel, B.~Thirion, O.~Grisel,
  M.~Blondel, P.~Prettenhofer, R.~Weiss, V.~Dubourg, J.~Vanderplas, A.~Passos,
  D.~Cournapeau, M.~Brucher, M.~Perrot, E.~Duchesnay,
  \href{https://dl.acm.org/doi/10.5555/1953048.2078195}{Scikit-learn: Machine
  {L}earning in {P}ython}, J. Mach. Learn. Res. 12 (2011) 2825–2830.
\newline\urlprefix\url{https://dl.acm.org/doi/10.5555/1953048.2078195}

\bibitem{OpenML2013}
J.~Vanschoren, J.~N. van Rijn, B.~Bischl, L.~Torgo,
  \href{https://doi.org/10.1145/2641190.2641198}{Open{ML}: Networked science in
  machine learning}, SIGKDD Explor. Newsl. 15~(2) (2014) 49–60.
\newblock \href {https://doi.org/10.1145/2641190.2641198}
  {\path{doi:10.1145/2641190.2641198}}.
\newline\urlprefix\url{https://doi.org/10.1145/2641190.2641198}

\bibitem{Dua2019UCI}
D.~Dua, C.~Graff, \href{http://archive.ics.uci.edu/ml}{{UCI} machine learning
  repository} (2017).
\newline\urlprefix\url{http://archive.ics.uci.edu/ml}

\bibitem{krizhevsky2009learning}
A.~Krizhevsky,
  \href{https://www.cs.toronto.edu/~kriz/learning-features-2009-TR.pdf}{Learning
  multiple layers of features from tiny images} (2009) 32--33.
\newline\urlprefix\url{https://www.cs.toronto.edu/~kriz/learning-features-2009-TR.pdf}

\bibitem{collobert2002parallel}
R.~Collobert, S.~Bengio, Y.~Bengio,
  \href{https://proceedings.neurips.cc/paper/2001/file/36ac8e558ac7690b6f44e2cb5ef93322-Paper.pdf}{A
  parallel mixture of {SVM}s for very large scale problems}, in: T.~Dietterich,
  S.~Becker, Z.~Ghahramani (Eds.), Advances in Neural Information Processing
  Systems, Vol.~14, MIT Press, 2001, pp. 633--640.
\newline\urlprefix\url{https://proceedings.neurips.cc/paper/2001/file/36ac8e558ac7690b6f44e2cb5ef93322-Paper.pdf}

\bibitem{alpaydin1998cascading}
E.~Alpaydin, C.~Kaynak,
  \href{http://www.kybernetika.cz/content/1998/4/369}{Cascading classifiers},
  Kybernetika 34~(4) (1998) 369--374.
\newline\urlprefix\url{http://www.kybernetika.cz/content/1998/4/369}

\bibitem{fisher1936use}
R.~A. Fisher,
  \href{https://onlinelibrary.wiley.com/doi/abs/10.1111/j.1469-1809.1936.tb02137.x}{The
  use of multiple measurements in taxonomic problems}, Annals of Eugenics 7~(2)
  (1936) 179--188.
\newline\urlprefix\url{https://onlinelibrary.wiley.com/doi/abs/10.1111/j.1469-1809.1936.tb02137.x}

\bibitem{hsu2002comparison}
C.-W. Hsu, C.-J. Lin, \href{https://doi.org/10.1109/72.991427}{A comparison of
  methods for multiclass support vector machines}, IEEE Transactions on Neural
  Networks 13~(2) (2002) 415--425.
\newblock \href {https://doi.org/10.1109/72.991427}
  {\path{doi:10.1109/72.991427}}.
\newline\urlprefix\url{https://doi.org/10.1109/72.991427}

\bibitem{higuera2015self}
C.~Higuera, K.~J. Gardiner, K.~J. Cios,
  \href{https://doi.org/10.1371/journal.pone.0129126}{Self-organizing feature
  maps identify proteins critical to learning in a mouse model of down
  syndrome}, PLOS ONE 10~(6) (2015) 1--28.
\newblock \href {https://doi.org/10.1371/journal.pone.0129126}
  {\path{doi:10.1371/journal.pone.0129126}}.
\newline\urlprefix\url{https://doi.org/10.1371/journal.pone.0129126}

\bibitem{strehl2002cluster}
A.~Strehl, J.~Ghosh, \href{https://doi.org/10.1162/153244303321897735}{Cluster
  ensembles --- a knowledge reuse framework for combining multiple partitions},
  J. Mach. Learn. Res. 3 (2003) 583–617.
\newblock \href {https://doi.org/10.1162/153244303321897735}
  {\path{doi:10.1162/153244303321897735}}.
\newline\urlprefix\url{https://doi.org/10.1162/153244303321897735}

\bibitem{2020SciPy-NMeth}
P.~Virtanen, R.~Gommers, T.~E. Oliphant, M.~Haberland, T.~Reddy, D.~Cournapeau,
  E.~Burovski, P.~Peterson, W.~Weckesser, J.~Bright, S.~J. van~der Walt,
  M.~Brett, J.~Wilson, K.~J. Millman, N.~Mayorov, A.~R.~J. Nelson, E.~Jones,
  R.~Kern, E.~Larson, C.~J. Carey, {\.{I}}.~Polat, Y.~Feng, E.~W. Moore,
  J.~VanderPlas, D.~Laxalde, J.~Perktold, R.~Cimrman, I.~Henriksen, E.~A.
  Quintero, C.~R. Harris, A.~M. Archibald, A.~H. Ribeiro, F.~Pedregosa, P.~van
  Mulbregt, A.~Vijaykumar, A.~P. Bardelli, A.~Rothberg, A.~Hilboll,
  A.~Kloeckner, A.~Scopatz, A.~Lee, A.~Rokem, C.~N. Woods, C.~Fulton,
  C.~Masson, C.~H{\"a}ggstr{\"o}m, C.~Fitzgerald, D.~A. Nicholson, D.~R. Hagen,
  D.~V. Pasechnik, E.~Olivetti, E.~Martin, E.~Wieser, F.~Silva, F.~Lenders,
  F.~Wilhelm, G.~Young, G.~A. Price, G.-L. Ingold, G.~E. Allen, G.~R. Lee,
  H.~Audren, I.~Probst, J.~P. Dietrich, J.~Silterra, J.~T. Webber,
  J.~Slavi{\v{c}}, J.~Nothman, J.~Buchner, J.~Kulick, J.~L. Sch{\"o}nberger,
  J.~V. de~Miranda~Cardoso, J.~Reimer, J.~Harrington, J.~L.~C. Rodr{\'i}guez,
  J.~Nunez-Iglesias, J.~Kuczynski, K.~Tritz, M.~Thoma, M.~Newville,
  M.~K{\"u}mmerer, M.~Bolingbroke, M.~Tartre, M.~Pak, N.~J. Smith, N.~Nowaczyk,
  N.~Shebanov, O.~Pavlyk, P.~A. Brodtkorb, P.~Lee, R.~T. McGibbon,
  R.~Feldbauer, S.~Lewis, S.~Tygier, S.~Sievert, S.~Vigna, S.~Peterson,
  S.~More, T.~Pudlik, T.~Oshima, T.~J. Pingel, T.~P. Robitaille, T.~Spura,
  T.~R. Jones, T.~Cera, T.~Leslie, T.~Zito, T.~Krauss, U.~Upadhyay, Y.~O.
  Halchenko, Y.~V{\'a}zquez-Baeza, S.~1.0~Contributors,
  \href{https://doi.org/10.1038/s41592-019-0686-2}{Scipy 1.0: fundamental
  algorithms for scientific computing in {P}ython}, Nature Methods 17~(3)
  (2020) 261--272.
\newblock \href {https://doi.org/10.1038/s41592-019-0686-2}
  {\path{doi:10.1038/s41592-019-0686-2}}.
\newline\urlprefix\url{https://doi.org/10.1038/s41592-019-0686-2}

\end{thebibliography}
\end{singlespace}





\end{document}